\pdfoutput=1

\documentclass[11pt]{article}

\usepackage[preprint]{acl}

\usepackage{times}
\usepackage{latexsym}

\usepackage[T1]{fontenc}

\usepackage[utf8]{inputenc}

\usepackage{microtype}

\usepackage{inconsolata}

\usepackage[T1]{fontenc}    %
\usepackage{hyperref}       %
\usepackage{url}            %
\usepackage{booktabs}       %
\usepackage{amsfonts}       %
\usepackage{nicefrac}       %
\usepackage{microtype}      %
\usepackage{longtable}
\usepackage{colortbl} 
\usepackage{wrapfig}
\usepackage{tcolorbox}
\usepackage{verbatim} %
\usepackage{fvextra} %
\usepackage{listings} %

\usepackage{floatflt}
\usepackage{subcaption} %
\usepackage{adjustbox} %
\usepackage{array}
\usepackage{graphicx}
\usepackage{enumitem}
\usepackage[toc,page]{appendix} %

\newtcolorbox[auto counter, number within=section]{customprompt}[2][]{%
    colback=gray!10,            %
    colframe=gray!80,           %
    coltitle=black,             %
    fonttitle=\bfseries,        %
    colbacktitle=gray!30,       %
    title={#2},                 %
    #1                          %
}
\usepackage{hyperref}
\usepackage{url}

\newcommand{\trainset}{\texttt{SWE-Fixer-Train-110K}}
\newcommand{\trainretrieval}{\texttt{SWE-Fixer-Retrieval-80K}}
\newcommand{\trainedit}{\texttt{SWE-Fixer-Editing-70K}}
\newcommand{\traineditcot}{\texttt{SWE-Fixer-Editing-70K-CoT}}
\newcommand{\trainsmall}{\texttt{SWE-Fixer-Train-10K}}

\definecolor{ForestGreen}{rgb}{0.133, 0.545, 0.133}

\title{SWE-Fixer: Training Open-Source LLMs for Effective and Efficient GitHub Issue Resolution}

\author{
\hspace{-0.5cm}
Chengxing Xie\thanks{Equal Contribution. 
}~~\textsuperscript{\rm 1, \rm 2} \quad 
Bowen Li\footnotemark[1]~~\textsuperscript{\rm 1} \quad 
Chang Gao\footnotemark[1]~~\textsuperscript{\rm 1, \rm 3} \\
\hspace{-0.5cm}
\textbf{He Du}\textsuperscript{\rm 1}\quad 
\textbf{Wai Lam}\textsuperscript{\rm 3}\quad 
\textbf{Difan Zou}\textsuperscript{\rm 4}\quad 
\textbf{Kai Chen}\textsuperscript{\rm 1}~\thanks{Corresponding author.} \\
\hspace{-0.6cm}
\textsuperscript{\rm 1}Shanghai AI Laboratory 
\quad \textsuperscript{\rm 2}Xidian University 
\quad \textsuperscript{\rm 3}The Chinese University of Hong Kong \\
\hspace{-0.8cm}
\textsuperscript{\rm 4}The University of Hong Kong \\
<\texttt{xiechengxing34@gmail.com}>~~<\texttt{libowen.ne@gmail.com}>~~<\texttt{chenkai@pjlab.org.cn}>
}

\begin{document}
\maketitle
\begin{abstract}

Large Language Models (LLMs) have demonstrated remarkable proficiency across a variety of complex tasks. One significant application of LLMs is in tackling software engineering challenges, particularly in resolving real-world tasks on GitHub by fixing code based on the issues reported by the users. However, many current approaches rely on proprietary LLMs, which limits reproducibility, accessibility, and transparency. The critical components of LLMs for addressing software engineering issues and how their capabilities can be effectively enhanced remain unclear. To address these challenges, we introduce SWE-Fixer, a novel open-source framework designed to effectively and efficiently resolve GitHub issues. SWE-Fixer comprises two essential modules: a code file retrieval module and a code editing module. The retrieval module employs BM25 along with a lightweight model to achieve coarse-to-fine file retrieval. Subsequently, the code editing module utilizes the other model to generate patches for the identified files. To mitigate the lack of publicly available datasets, we compile an extensive dataset that includes 110K GitHub issues along with their corresponding patches and train the two models of SWE-Fixer separately.  
We assess our approach on the SWE-Bench Lite and Verified benchmarks, achieving competitive performance among open-source models with scores of 22.0\% and 30.2\%. Furthermore, SWE-Fixer reaches state-of-the-art performance (24.7\% on Lite and 32.8\% on Verified) with PASS\_TO\_PASS (P2P) filtering. 
Additionally, our approach requires only two model calls per instance, making it significantly more efficient than existing methods. These results highlight the effectiveness of SWE-Fixer in real-world code-fixing scenarios.
We will make our model, dataset, and code publicly available at \url{https://github.com/InternLM/SWE-Fixer}.

\end{abstract}

\section{Introduction}

Large Language Models (LLMs) have demonstrated remarkable progress in code-related tasks, particularly excelling in code generation benchmarks such as HumanEval~\citep{chen2021codex_humaneval}
and LiveCodeBench~\citep{jain2024livecodebench}. 
However, these benchmarks primarily focus on single-file scenarios with constrained context scope, failing to capture the complexity and interdependencies inherent in real-world software development.
To bridge this gap,
researchers introduce SWE-Bench~\citep{jimenez2023swebench}, a benchmark designed to assess LLMs' ability to resolve real-world GitHub issues by generating code patches and verifying their correctness through issue-specific test cases.

Current approaches to resolving the Github issues in SWE-Bench can be broadly categorized into two main paradigms: \emph{agent}~\citep{wang2024opendevin,ma2024lingma,moatless} and \emph{pipeline}~\citep{xia2024agentless,appmap2024navie,li2025patchpilot}. Agent-based systems rely on LLMs dynamically determining the next action, allowing them to autonomously explore a codebase and resolve issues. 
In contrast, pipeline-based methods guide LLMs through a series of predefined steps, such as initially identifying the defective files and subsequently editing them to resolve the issue. 
Despite significant advancements in agent and pipeline-based frameworks, most existing solutions depend on powerful proprietary models such as GPT-4o and Claude-3.5-Sonnet. While these solutions are effective, they are often costly and lack transparency, preventing a deeper understanding and further improvements in the problem-solving ability of LLMs. 
This motivates us to develop open-source LLMs to resolve the Github issues effectively and efficiently.

Along with these two paradigms, their corresponding training approaches are also largely different and face unique challenges. In particular, training open-source models in an agent-based framework presents several challenges. First, compared with frontier closed-source models such as Claude-3.5-Sonnet, existing open-source models lack the robust agent capabilities necessary for self-directed decision-making, long-term planning, and effective tool use, which may limit their performance on complex tasks.
Second, constructing training data for agent-based methods often requires access to a real execution environment, which can be difficult to set up.
Even with such an environment, the performance of the most advanced models is still not satisfactory to resolve real-world issues, making trajectory collection both expensive and inefficient.
In contrast, while the pipeline-based approach could be less flexible than the agent-based approach, it simplifies data construction and model training by explicitly defining each subtask.
Notably, the Agentless framework \citep{xia2024agentless} employs a complex, multi-stage design that organizes problem-solving procedures into a pipeline. This approach has demonstrated competitive performance on the GitHub issue resolution task using proprietary models. For instance, the framework identifies problematic code through a multi-step process: LLMs first filter candidate files, then apply both LLM-based and dense retrieval on the selected subset, followed by LLM-guided localization of relevant classes, functions, and code lines. 
While this pipeline provides a structured training framework, the primary bottleneck lies in the scarcity of high-quality real-world training data for all intermediate tasks, which hinders the practical training of open-source models capable of accurately executing each step in the pipeline.

\begin{figure*}[t]
    \centering
    \includegraphics[width=0.9\linewidth]{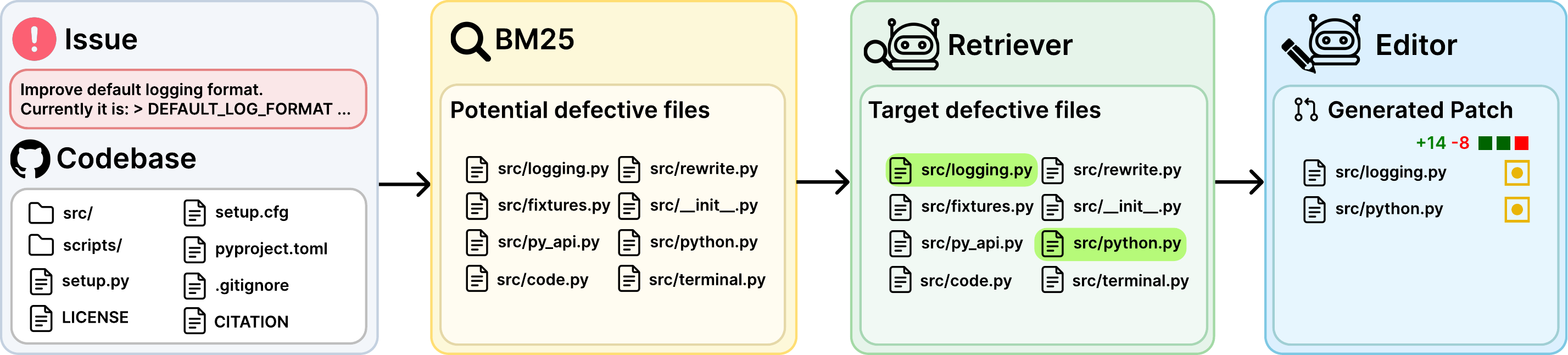} 
    \caption{The pipeline of SWE-Fixer. Our framework contains two components: code file retrieval and code editing. The two frames of BM25 and Retriever indicate our \textit{coarse-to-fine} retrieval method. The editor frame represents the code editing task. 
    }
    \label{fig:workflow}
\end{figure*}

To address these challenges, we introduce SWE-Fixer, a simple yet effective pipeline-based approach for training open-source models to resolve Github issues.
Unlike Agentless~\citep{xia2024agentless}, which employs a complex pipeline, SWE-Fixer streamlines the process by reducing the number of reasoning steps, facilitating more effective model training, and significantly lowering inference costs.
Our method decomposes the task into two core subtasks: code file retrieval and code editing (see Figure \ref{fig:workflow}).
For the retrieval task, we use a \emph{coarse-to-fine} strategy combining BM25 for initial file retrieval and a model to identify the defective files from the BM25 results.
Once the defective files are identified, the editing model generates a code patch to resolve the issue, trained using chain-of-thought data.
The streamlined pipeline design facilitates easier training data construction and efficient inference, eliminating the need for complex manipulations.
We curate a large-scale dataset comprising 110K instances for the training of both retrieval and editing tasks.
To ensure data quality, we apply rigorous filtering techniques, making the training set both extensive and reliable.
For model implementation, we fine-tune the Qwen2.5~\citep{qwen2024qwen25technicalreport} base series models, which include a 7B retriever and a 72B editor.

SWE-Fixer achieves 22.0\% on SWE-Bench Lite and 30.2\% on SWE-Bench Verified, performing comparably to existing SOTA open-source models. Additionally, with P2P filtering, SWE-Fixer achieves Best@1 performance of 24.7\% on SWE-Bench Lite and 32.8\% on SWE-Bench Verified. These results set a new SOTA for open-source model-based approaches.
Furthermore, SWE-Fixer requires \emph{only two model calls per instance}, making it significantly more efficient than existing open-source frameworks while maintaining superior performance.
Notably, compared to frameworks built on proprietary models, SWE-Fixer outperforms several methods on both benchmarks, particularly those based on GPT-4, GPT-4o, and Claude-3-Opus, demonstrating exceptional efficiency and effectiveness.
We also conduct a comprehensive study on training configurations for both tasks
to provide insights into optimizing future systems.
Additionally, the trained retriever and editor from SWE-Fixer have the potential to serve as modular components in agent-based systems, further enhancing the efficiency and effectiveness of existing agent-based approaches. 
Our contributions can be summarized as follows:
\begin{itemize}[leftmargin=0.4cm]
    \item \textbf{State-of-the-art performance:} We propose a novel pipeline-based method that leverages open-source models, achieving state-of-the-art Best@1 performance on SWE-Bench Lite and Verified with P2P filtering compared to other open-source model-based approaches.
    \item \textbf{Large-scale dataset curation:} We curate a large-scale training dataset comprising 110K instances with rigorous filtering techniques to ensure both the extensiveness and reliability.
    \item \textbf{Comprehensive Analysis:} We provide an in-depth analysis of the data configurations for each subtask, enabling optimized task performance. 
\end{itemize}

\section{Related Works}

\paragraph{Code Large Language Models}

The use of large language models (LLMs) for coding tasks has garnered substantial attention recently. Numerous LLMs, trained on extensive datasets of code, have been introduced, including closed-source models such as GPT-4~\citep{gpt4}, Gemini~\citep{team2023gemini}, and Claude-3.5~\citep{claude3_5} and open-source models such as Mixtral~\citep{jiang2024mixtral}, Deepseek-Coder~\cite{zhu2024deepseekcoderv2}, and Qwen-Coder~\citep{hui2024qwen2}. Beyond these general-purpose code models, significant efforts have been made to tackle specific software engineering challenges with specialized LLMs. For instance, several studies have introduced strategies to improve program generation \citep{wang2024planning,huang2023enhancing,zheng2023self,chen2024jumpcoder,liu2024your} or enhance program repair \citep{lin2024one,jiang2023impact,xia2022practical}. 
In the domain of code retrieval, researchers have developed advanced techniques for refining code representations \citep{bui2021self,li2024rewriting,martinez2024improving,saieva2023reinforest} and improving fault retrieval \citep{qin2024agentfl,du2024generalization}.

\paragraph{LLMs for GitHub Issue Solving}

Resolving repository issues with LLMs has gained attention, with SWE-Bench~\citep{jimenez2023swebench} emerging as a key benchmark, featuring 2,294 issues from 12 high-quality GitHub repositories. 
Various agent-based approaches, including SWE-agent~\citep{yang2024sweagent}, Autocoderover~\citep{zhang2024autocoderover}, OpenHands~\citep{wang2024opendevin}, and Moatless Tools~\citep{moatless}, leverage proprietary models and tools for autonomous code exploration and issue resolution.
SWE-SynInfer~\citep{ma2024lingma} and SWE-Gym~\citep{pan2024swegym} adopt agent-based frameworks and explore open-source model training on this task.
Additionally, Agentless~\citep{xia2024agentless}, a pipeline-based approach, demonstrates competitive performance against agent-based methods.
Agentless shares a philosophy similar to SWE-Fixer but differs significantly in its approach. 
Concretely, Agentless features a highly sophisticated architectural design on both code retrieval and editing stages that leverages powerful proprietary models, making it challenging to adapt for training open-source models (as shown in our experiments in Section \ref{sec:ablation_study}).
In contrast, SWE-Fixer offers a simpler and more robust design, facilitating easier training data construction, making it better suited for finetuning open-source models, and delivering strong results on SWE-Bench.

\begin{figure*}[ht]
    \centering
    \begin{subfigure}{\textwidth}
        \centering
        \includegraphics[width=0.9\textwidth]{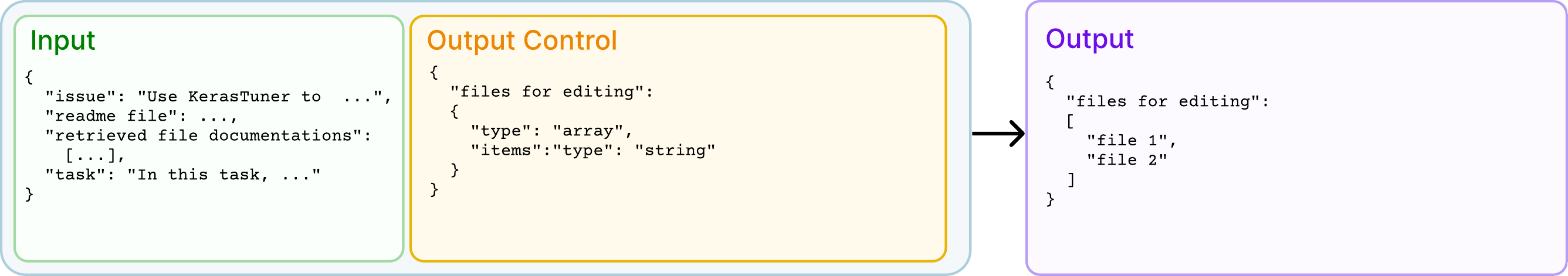} %
        \caption{The code file retrieval task.}
        \label{fig:}
    \end{subfigure}   
    \begin{subfigure}{\textwidth}
        \centering
        \includegraphics[width=0.9\textwidth]{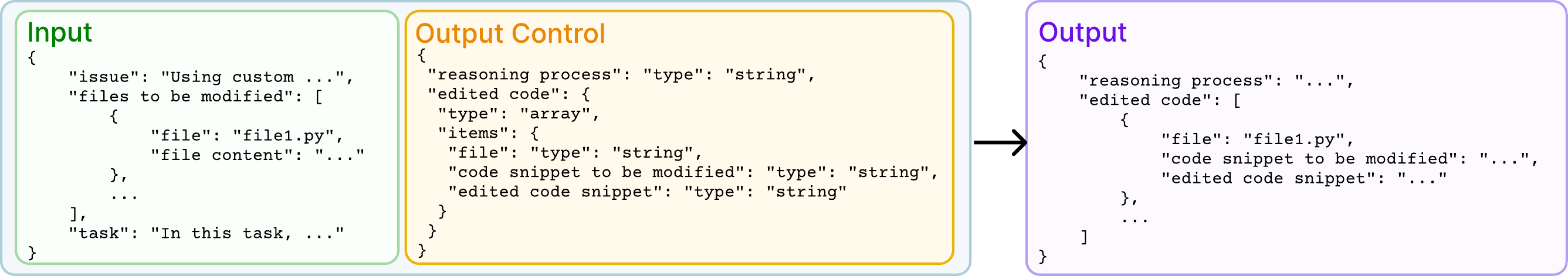} %
        \caption{The code editing task.}
        \label{fig:edit_input_horizon_original}
    \end{subfigure}
    \caption{Structured representation of the retrieval/editing task's inputs and outputs.}
    \label{fig:input_output_structure}
\end{figure*}

\section{SWE-Fixer}

\subsection{Overview}

The SWE-Fixer framework is designed to efficiently address real-world GitHub issues by generating code patches that specify the necessary modifications to a repository’s codebase. We divide this task into two subtasks: code file retrieval and code editing.
Identifying the correct files for modification is a major challenge, as repositories typically contain a vast number of files. To tackle this, we introduce a coarse-to-fine retrieval approach that efficiently narrows down the search space and accurately locates the files requiring edits.
Once the files are identified, a specialized editor model is employed to produce high-quality code patches that effectively resolve the reported issues.
As shown in Figure \ref{fig:workflow}, we adopt a structured pipeline approach to systematically optimize open-source models for each subtask. The following sections provide a detailed breakdown of these components within the SWE-Fixer framework.

\subsection{Code File Retrieval}

To efficiently identify relevant files for modification in a repository, we adopt a \emph{coarse-to-fine} strategy.
First, as seen in the BM25 part of Figure \ref{fig:workflow}, we use BM25~\citep{robertson2009probabilistic} to retrieve the 30 most relevant files to the issue, treating the issue description as a query and the code files as documents.
Building on these initial retrieval results, we then finetune a retriever model to further refine the selection and accurately identify which files require modification.
We chose BM25 over the dense retrieval method used in the Agentless~\citep{xia2024agentless} and Moatless Tools ~\citep{moatless} because it provides a lightweight, scalable and robust approach for initial file retrieval, especially when the number of files in a repository is large.
BM25 provides an efficient and effective initial retrieval mechanism, which is then complemented by a language model refinement step to narrow down the selection to the most relevant files.

Given the large potential context of all files, we take inspiration from the Agentless framework and use file documentations (skeletons) as input.
A file documentation includes module docstrings, class headers, class methods, and function signatures, retaining only the signatures of class methods and the first and last five lines of functions (see example in Figure~\ref{fig:file_skeleton}). 
This approach significantly reduces context size while preserving the essential information for effective file-level retrieval.

\subsection{Code Editing}

The code editing task involves generating a patch to resolve an issue based on the relevant files. 
Previous works on model training primarily adopt agentic approaches without explicitly treating code editing as a standalone sub-task. 
In this work, we identify code editing as the main bottleneck in the entire workflow and explicitly focus on improving its effectiveness.
During training, we utilize gold defective files, i.e., those within the patches. For inference, we use retrieved files.
To provide sufficient context, we include the content of all retrieved files, even though only a small portion of the content in these files would be modified.
Including additional context helps the model better understand the issue and apply the necessary changes effectively.

To assist the model in generating valid code modifications, we define a structured output that includes three key components (see Appendix \ref{sec:our_edit_output_example}).
The first component is the file path, which specifies the location of the file to be modified.
The second is the original code block, which includes the specific code snippet to be edited along with its line numbers.
The third component is the modified code block, which represents the final modification result but excludes line numbers as the new line numbers are difficult for the model to calculate.
The input includes complete file content with line numbers.
This approach ensures that the model can efficiently identify the lines to edit and produce valid modifications.
In contrast, the standard patch format\footnote{https://git-scm.com/docs/diff-generate-patch} requires line number calculations in the output hunk, which adds complexity.
Our structured output simplifies training and can be transformed to patches automatically for evaluation.

\section{Model Training}

\subsection{Structured Instruction Tuning}

We adopt a structured approach, JsonTuning~\citep{gao2023jsontuning}, to train our models, enhancing the overall pipeline performance.
As shown in Figure \ref{fig:input_output_structure}, the input JSON structure includes task input elements, task instructions, and output control information, while the output JSON structure consists of task output elements.
For the code file retrieval task, input elements include issues and file documentations, and output elements include files for editing.
For the code editing task, input elements include issues and file content, and output elements include reasoning processes, original code snippets, and modified code snippets.
During training, we provide the model with a task-specific input in JSON format and expect it to generate a corresponding JSON-formatted output. 
The use of JsonTuning offers several advantages: (1) By incorporating explicit task structures into the input and output representations during instruction tuning, JsonTuning enhances the model’s comprehension of critical task elements and their interrelations, thereby improving generalization capabilities. (2) Given the inherently structured nature of code, JsonTuning efficiently leverages this structured information. (3) JsonTuning provides a structured output that is more robust compared to generating patches, which often involve stricter syntax and pose greater challenges for model interpretation. 
To further improve performance, we apply a corresponding post-processing procedure (see Appendix \ref{sec:post_pocessing} for more details).

\subsection{Chain-of-Thought Data Construction}

For tasks that require strong reasoning capabilities, such as math or coding, LLMs can improve their performance by using Chain-of-Thought.
However, the data collected from real-world scenarios typically includes only the specific codebase associated with an issue and the corresponding gold patches, without capturing the intermediate reasoning behind these patches.
To address this gap, we construct Chain-of-Thought (CoT) data for the code editing task.
A straightforward approach to generating CoT data involves prompting a teacher model to produce the reasoning process and then performing rejection sampling.
However, this approach presents several challenges.
Due to the complexity of the task, even advanced proprietary models struggle to achieve high accuracy when performing code editing independently.
Additionally, without access to a real code execution environment, we are unable to verify the patches generated by the teacher model through execution, making standard rejection sampling infeasible.

To address these challenges, we adopt a methodology inspired by the \emph{rationalization} approach in \citet{zelikman2022star}.
Specifically, we use the gold patches as part of the input to guide the teacher model in generating both the reasoning process and the corresponding patches.
The model is tasked with producing a reasoning chain and code patch as if it were \emph{unaware} of the correct answer. 
Detailed prompts can be found in Appendix \ref{sec:cot_generation_prompt}.
We use GPT-4o for the generation, and the resulting reasoning chains are generally coherent and sound.

\begin{table*}[t]
\caption{Performance of various models/methods on SWE-Bench Lite and Verified. $^\Diamond$: Finetuned open-source models. $^\ast$: All results are reported as Pass@1 or Best@1 in this table, except for SWE-Gym, which uses a specially trained 32B verifier. P2P filtering means that we use the P2P tests in the repository to discard patches that break them, retaining only those that pass all P2P tests.
}
\label{tab:main_result}

\centering

\scalebox{0.73}{
\begin{tabular}{@{}llccc@{}}
\toprule
\textbf{Method}         & \textbf{Model}                & \textbf{Type} & \textbf{Verified} & \textbf{Lite} \\

\midrule
\multicolumn{5}{c}{\cellcolor{gray!11}{\textbf{\emph{Open-source Methods w/ Proprietary Models}}}} \\
\midrule
RAG~\citep{jimenez2023swebench}                    & GPT-4                       & Pipeline & 2.8            & 2.7        \\
RAG~\citep{jimenez2023swebench}                    & Claude-3-Opus              & Pipeline & 7.0            & 4.3        \\

SWE-agent~\citep{yang2024sweagent}              & Claude-3-Opus              & Agent & 18.2           & 11.7       \\
SWE-agent~\citep{yang2024sweagent}              & GPT-4o                      & Agent & 23.0           & 18.3       \\
AppMap Navie~\citep{appmap2024navie}           & GPT-4o                      & Pipeline & 26.2           & 21.7       \\
AutoCodeRover~\citep{zhang2024autocoderover}          & GPT-4                       & Agent & -           & 19.0       \\
SWE-agent + RepoGraph~\citep{ouyang2024repograph}    & GPT-4o                      & Agent & -           & 20.3       \\

AutoCodeRover + RepoGraph~\citep{ouyang2024repograph}  & GPT-4                       & Agent & -           & 21.3       \\
OpenHands~\citep{wang2024opendevin}          & GPT-4o                      & Agent & -          & 22.0       \\

AutoCodeRover~\citep{zhang2024autocoderover}          & GPT-4o                      & Agent & 28.8           & 22.0       \\

SWE-SynInfer~\citep{ma2024lingma}            & GPT-4o                      & Agent & 31.8           & 20.7       \\
SWE-agent~\citep{yang2024sweagent}              & Claude-3.5-Sonnet          & Agent & 33.6           & 23.0       \\
Agentless~\citep{xia2024agentless}              & GPT-4o                      & Pipeline & 38.8          & 32.0       \\
Moatless Tools~\citep{moatless}          &  Claude-3.5-Sonnet-\texttt{20241022}                     & Agent & -           & 38.3       \\
AutoCodeRover-v2.0~\citep{zhang2024autocoderover}        &  Claude-3.5-Sonnet-\texttt{20241022}          & Agent & -           & 46.2       \\

Agentless~\citep{xia2024agentless}              & Claude-3.5-Sonnet-\texttt{20241022}             & Pipeline & 50.8         & 40.7       \\
OpenHands~\citep{wang2024opendevin}          & Claude-3.5-Sonnet-\texttt{20241022}                      & Agent & 53.0          & 41.7       \\

\midrule
\multicolumn{5}{c}{\cellcolor{gray!11}{\emph{\textbf{Open-source Methods w/ Open-source Models}}}} \\
\midrule

RAG~\citep{jimenez2023swebench}                    & SWE-Llama-13B$^\Diamond$         & Pipeline & 1.2            & 1.0        \\
AutoCodeRover~\citep{liu2024codexgraph}          & Qwen2-72B-Instruct         & Agent & -            & 9.3             \\
SWE-Gym (Best@1)~\citep{pan2024swegym}          & SWE-Gym-32B$^\Diamond$         & Agent & 20.6            & 15.3             \\

SWE-SynInfer~\citep{ma2024lingma}            & Lingma-SWE-GPT-72B$^\Diamond$         & Agent & 30.2           & 22.0       \\
SWE-Gym (Best@8 w/ Verifier)$^\ast$~\citep{pan2024swegym}          & SWE-Gym-32B$^\Diamond$         & Agent & 29.8            & \textbf{26.0}             \\
SWE-Search~\citep{antoniades2024swe_search} & Qwen2.5-72b-Instruct  & Agent & - &   24.7  \\
\textbf{SWE-Fixer}   & SWE-Fixer-72B$^\Diamond$                     & Pipeline & 30.2          & 22.0             \\
\textbf{SWE-Fixer + P2P Filtering}   & SWE-Fixer-72B$^\Diamond$                     & Pipeline & \textbf{32.8}          & 24.7             \\
\bottomrule

\end{tabular}
}

\end{table*}

\section{Experiments}

\subsection{Experimental Setup}

We finetune the Qwen2.5~\citep{qwen2024qwen25technicalreport} base series models to implement SWE-Fixer, which consists of a 7B code retriever and a 72B code editor. 
The training is conducted on 96 Nvidia A800 GPUs using the xtuner-lite framework~\citep{2023xtuner}, with a global batch size of 96 and a 64K context window.
We evaluate SWE-Fixer on two datasets: SWE-Bench Lite and SWE-Bench Verified. SWE-Bench Lite, part of the official SWE-Bench benchmark~\citep{jimenez2023swebench}, consists of 300 selected instances optimized for efficient evaluation. 
SWE-Bench Verified, proposed by OpenAI, is a human-validated subset that offers a more reliable evaluation. 
Each instance includes a real-world GitHub issue description paired with its corresponding codebase. 
Model-generated code patches are assessed using developer-written unit tests, and accuracy is calculated as the percentage of instances successfully resolved.
Some instances include PASS\_TO\_PASS (P2P) tests, which can be used to filter patches and ensure they do not affect unrelated functionality in the repository. This filtering is optional, and further details are provided in Appendix \ref{sec:p2p}.

\subsection{Dataset Preparation}

We construct our training dataset from \trainset. The detailed training data collection process is in Appendix \ref{sec:trainging_data_collection}.
For the code file retrieval task, we filter out those where the gold defective files do not appear in the top 30 retrieved files, or where the total length exceeds the 64K context window limit. 
This results in 80K valid training instances for the retrieval task, denoted as \trainretrieval.

For the code editing task, due to computational resource constraints and API limits, we sample a subset of 70K instances, referred to as \trainedit, and construct CoT data for this subset, \traineditcot.
To explore optimal data formats and training strategies, we further sample 200 repositories, each contributing 50 instances (10K instances total), to create a smaller dataset, \trainsmall, for the ablation study (see Section \ref{sec:ablation_study}).

\begin{table*}[t]
\centering
\caption{Performance comparison of different methods with open-source models on SWE-Bench Verified and Lite. `\#Model Calls' represents the number of model calls required per instance. 
$^\dag$: The minimum number of model calls needed. 
$^\ddag$: The average number of model calls needed.
$^*$: See Appendix \ref{sec:swe_search_step_estimation} for the detailed calculation.}

\scalebox{0.9}{
\begin{tabular}{l c c c c}
\toprule
\textbf{Method} & \textbf{Model} & \textbf{Verified} & \textbf{Lite} & \textbf{\#Model Calls} \\
\midrule
RAG~\citep{jimenez2023swebench} & SWE-Llama-13B & 1.2 & 1.0 & 1 \\
AutoCodeRover~\citep{liu2024codexgraph} & Qwen2-72B-Instruct & - & 9.3 & 4$^\dag$ \\
SWE-Gym (Best@1)~\citep{pan2024swegym} & SWE-Gym-32B & 20.6 & 15.3 & 29$^\ddag$ \\
SWE-SynInfer~\citep{ma2024lingma} & Lingma-SWE-GPT-72B & 30.2 & 22.0 & 6$^\dag$ \\
SWE-Search~\citep{antoniades2024swe_search} & Qwen2.5-72b-Instruct & - &  24.7 & 200$^*$ \\
\textbf{SWE-Fixer}   & SWE-Fixer-72B     & 30.2        & 22.0     & 2$^\dag$  \\
\textbf{SWE-Fixer + P2P Filtering} & SWE-Fixer-72B & \textbf{32.8} & \textbf{24.7} & 2$^\dag$ \\
\bottomrule
\end{tabular}
\label{tab:step-results}

}
\end{table*}
\subsection{Main Results}

\begin{table*}[t]
\centering
\small
\caption{Ablation study of the code file retrieval task on SWE-Bench Lite. The table compares precision and recall across different methods and training datasets, using Qwen2.5-7B as the base model. $^\ast$: The base setting utilizes a 64K context window, which includes the readme file and BM25-retrieved file documentation limited to 30 files. 
$^\ddag$: Trains with additional 100K editing data without CoT, which is sampled from \trainset.}
\scalebox{0.96}{
\begin{tabular}{lccc}
\toprule
\textbf{Method} & \textbf{Training Dataset} & \textbf{Precision(\%)} & \textbf{Recall(\%)} \\

\midrule
\multicolumn{4}{c}{\cellcolor{gray!11}{\textbf{\emph{BM25 Baseline}}}} \\
\midrule

BM25 Top-3 & - &18.9 &56.7 \\
BM25 Top-30 & - & 2.9& 86.7\\

\midrule
\multicolumn{4}{c}{\cellcolor{gray!11}{\textbf{\emph{Training on Limited Data}}}} \\
\midrule

Base setting$^\ast$       & \textbf{\trainsmall} & 65.4          & 67.3         \\
~~{\small - Remove readme}              & \trainsmall          & 65.2~\textcolor{red}{($\downarrow$~0.2)}  & 66.7~\textcolor{red}{($\downarrow$~0.6)} \\
~~{\small - 32K context limit}            & \trainsmall          & 64.3~\textcolor{red}{($\downarrow$~1.1)}  & 64.7~\textcolor{red}{($\downarrow$~2.6)} \\
~~{\small - Add file content}           & \trainsmall          & 59.7~\textcolor{red}{($\downarrow$~5.7)}  & 60.3~\textcolor{red}{($\downarrow$~7.0)} \\

\midrule
\multicolumn{4}{c}{\cellcolor{gray!11}{\textbf{\emph{Training on Full Data}}}} \\
\midrule

Base setting$^\ast$               & \trainretrieval            & 68.5~\textcolor{ForestGreen}{($\uparrow$~3.1)}  & 69.0~\textcolor{ForestGreen}{($\uparrow$~1.7)} \\
Default setting$^\ddag$          &  \trainretrieval ~\texttt{+ 100K Editing Data}     & \textbf{69.4}~\textcolor{ForestGreen}{($\uparrow$~4.0)}  & \textbf{69.7}~\textcolor{ForestGreen}{($\uparrow$~2.4)} \\ 
\bottomrule
\end{tabular}
}
\label{tab:ablation_retrieval}
\end{table*}
Table \ref{tab:main_result} presents the primary results on SWE-Bench Lite and SWE-Bench Verified. 
We categorize the results into two groups: methods based on proprietary models and those based on open-source models.\footnote{Closed-source methods are omitted from Table \ref{tab:main_result}, as the lack of technical transparency limits the discussion of meaningful insights.}

We begin by comparing SWE-Fixer with frameworks based on proprietary models. 
Remarkably, our approach outperforms several methods across both SWE-Bench Lite and Verified, especially those based on GPT-4, GPT-4o, and Claude-3-Opus. 
The exception is only Agentless~\citep{xia2024agentless} using GPT-4o. 
This is encouraging, as these baselines typically involve complex agent frameworks, specifically designed for the SWE-Bench task, and rely on advanced proprietary models. 
Our results suggest that SWE-Fixer offers a promising, cost-effective alternative to proprietary model-based frameworks.
We also note that Claude-3.5-Sonnet excels in coding tasks, delivering consistent improvements over all open-source methods, where our approach still lags behind such systems.

Our SWE-Fixer framework achieves comparable performance among the approaches based on open-source models. 
While SWE-Gym-32B~\citep{pan2024swegym} attains a higher Best@8 score on SWE-Bench Lite with its specially trained 32B verifier, it still falls behind SWE-Fixer on SWE-Bench Verified. 
SWE-Gym represents a promising direction by leveraging reinforcement learning in executable environments. 
However, its heavy reliance on human effort for environment construction limits its scalability.
SWE-Fixer achieves the same performance as SWE-Synifer, an agent-based approach that requires significantly more model calls and is much more complex.
With P2P tests filtering, SWE-Fixer achieves the highest Best@1 performance among all open-source model-based methods.

Beyond effectiveness, our method also demonstrates exceptional efficiency. 
As shown in Table \ref{tab:step-results}, apart from the RAG method, which performs poorly, SWE-Fixer requires only two model calls (7B retrieval + 72B editing) per instance, making it significantly more efficient than other methods while maintaining comparable performance. 
Notably, SWE-Search requires at least 200 model calls per instance yet achieves only a slightly higher score than SWE-Fixer on SWE-Bench Lite without P2P filtering, rendering it prohibitively expensive by comparison.

\subsection{Ablation Study}
\label{sec:ablation_study}
\subsubsection{Code File Retrieval}

As shown in Table \ref{tab:ablation_retrieval}, we conduct a series of ablation experiments on the code file retrieval task to evaluate the impact of various input configurations on model performance. The base setup employs a 64K context window that includes the readme file along with BM25-retrieved file documentation, limited to a maximum of 30 files. 
The model is trained to predict a list of files requiring modification.

\textbf{Irrelevant information adversely affects performance.}
Providing a relevant input context is critical in the code file retrieval task. 
Including unnecessary details, such as the entire file content, leads to a decline in performance since the retrieval task does not require such fine-grained code information.
Conversely, incorporating relevant information, such as the readme file, improves model performance.

\textbf{Smaller context windows reduce effectiveness.}
A smaller context window (32K) restricts the number of input files, which may lead to the exclusion of target defective files. This limitation lowers recall within the input context, ultimately resulting in poorer overall performance compared to a model using a larger 64K context window.

\textbf{Larger datasets improve model performance.}
Expanding the training dataset size from 10K to 80K significantly boosts the model's performance. Additionally, incorporating data from both the retrieval and editing tasks during training further enhances retrieval performance. This improvement highlights the beneficial influence of editing task data on the retrieval task.
The editing task may provide a deeper understanding of the relationship between the issue and the code, which improves the model's ability to retrieve relevant code.

\subsubsection{Code Editing}
\label{sec:code editing}

\begin{table}[h!]
\centering
\caption{Ablation study of the code editing task on SWE-Bench Lite with gold defective code files as input using Qwen2.5-72B and the \trainsmall. \emph{Cls\&Func} refers to classes and functions. $^\ast$: The default setting uses only file content with line numbers. $^\dag$: Contains only Cls\&Func content, without file content.
}
\scalebox{0.95}{
\begin{tabular*}{0.95\linewidth}{@{\extracolsep{\fill}}lc}
\toprule
\textbf{Method} & \textbf{Resolve (\%)} \\ 
\midrule
Default setting$^\ast$  & 20.0 \\
~~{ - Only Cls\&Func Content$^\dag$}  & 18.0~\textcolor{red}{($\downarrow$~2.0)} \\
~~{ - Add readme}  & 19.0~\textcolor{red}{($\downarrow$~1.0)} \\
~~{ - Remove Line Number}  & 14.0~\textcolor{red}{($\downarrow$~6.0)} \\
\bottomrule
\end{tabular*}
}
\label{tab:edit_ablation}
\end{table}

As shown in Table \ref{tab:edit_ablation}, we also conduct a detailed ablation study on the code editing task to evaluate the impact of different data configurations. All experiments use gold input (oracle defective files). The default configuration includes complete file content with line numbers, providing better contextual understanding and precise localization.

\textbf{Enhanced location information improves performance.}
Including line numbers acts as an anchor, helping LLMs locate and edit specific code snippets more effectively, which improves performance. Additional experiments on more fine-grained settings can be found in Appendix \ref{sec:more_finegained_exp}.

\textbf{Redundant or insufficient information reduces performance.}
The readme file, while useful in retrieval tasks, introduces high-level abstract information that is irrelevant to the editing task. Editing tasks require detailed understanding of flawed code, and the inclusion of readme information detracts from this, resulting in lower performance. 
Insufficient information, such as input limited to class and function content, also negatively impacts performance.
This suggests that class and function-only inputs omit essential context needed for effective editing.

\subsection{Scaling Trends of the Code Editing Task}

\begin{figure}[h]
    \centering
    \includegraphics[width=\linewidth]{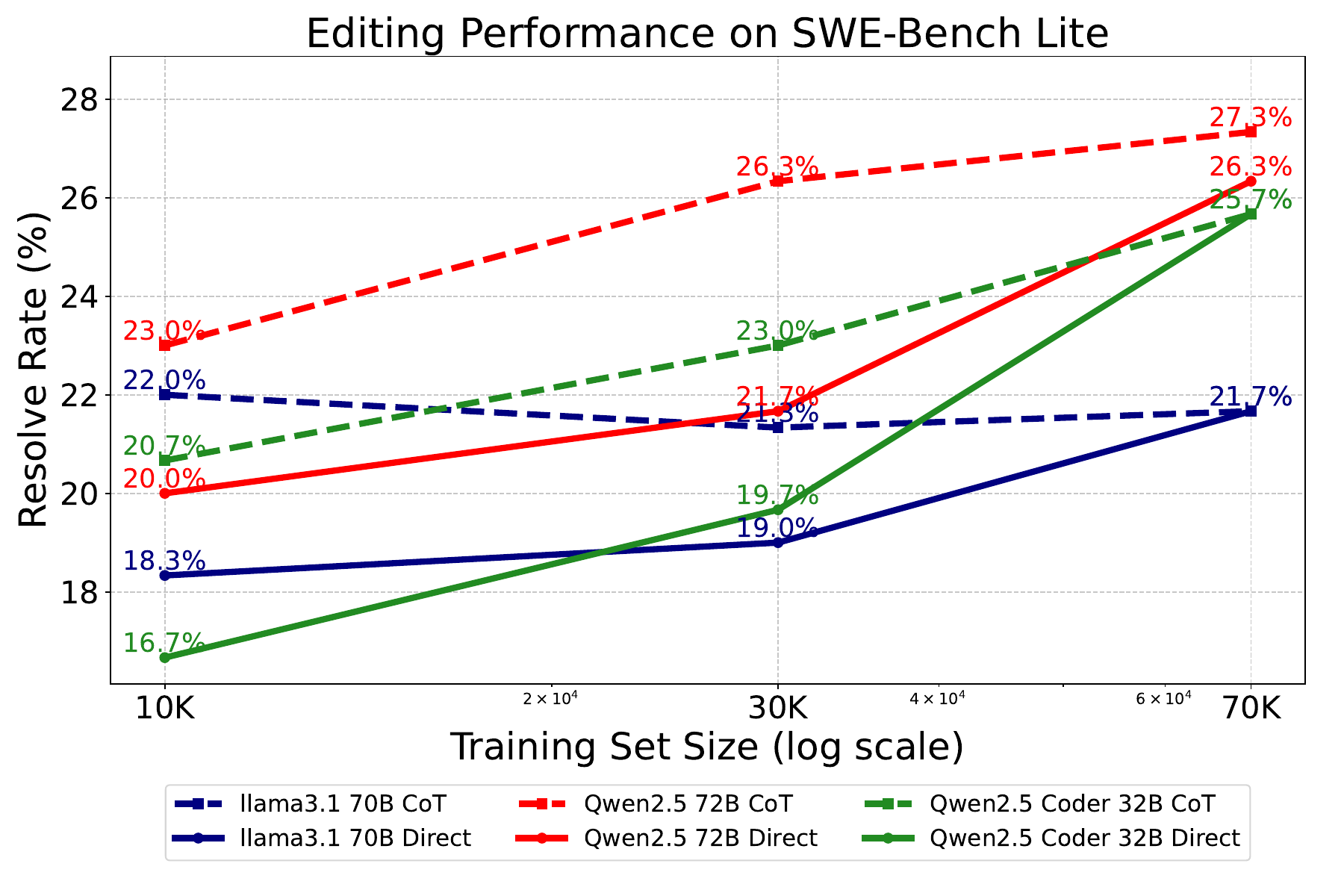}
    \caption{Performance of the editing task on SWE-Bench Lite across various models and training settings with gold defective code files as input. The x-axis is in logarithmic scale. Solid lines indicate methods that generate modified code directly, while dashed lines represent methods that incorporate a reasoning step before generating the modified code.}
    \label{fig:scaling_editing}
\end{figure}

Compared to the retrieval task, LLMs perform significantly worse on the editing task, making it the primary bottleneck in overall task performance. To further investigate LLMs' capabilities in code editing, we analyze their performance by training on datasets of varying scales and using gold input to test their editing ability on SWE-Bench Lite, as shown in Figure \ref{fig:scaling_editing}.
The direct training method demonstrates consistent performance improvements across models as the training data size increases. 
Stronger models exhibit steeper scaling trends compared to weaker models. 
For instance, while Qwen2.5-Coder-32B initially lags behind Llama3.1-70B with 10K training instances, it gradually surpasses Llama3.1-70B as more training data is available.
Notably, even at 70K instances, the performance curves for these models have not shown clear signs of plateauing, suggesting that increasing the training data size may still have the potential to yield additional performance gains for direct training.
In contrast, CoT training exhibits varying scaling trends across models.
For Qwen2.5-Coder-32B and Qwen2.5-72B, the CoT method demonstrates a consistent upward trend in performance, significantly outperforming direct training.
However, for Llama-3.1-70B, which may lack sufficient capacity in this domain, increasing the CoT training data does not lead to sustained performance improvements. 
Additionally, as the training dataset grows, the performance gap between CoT and direct training narrows.

\section{Conclusion}

In this paper, we present SWE-Fixer, an open-source framework designed to efficiently and effectively address real-world software engineering challenges using finetuned open-source models.
SWE-Fixer adopts a pipeline-based approach, dividing the task into two subtasks: code file retrieval and code editing, requiring only two steps to generate the final results.
To support model training, we curate a large-scale, real-world dataset and construct task-specific training data for both subtasks.
Our method demonstrates impressive performance on SWE-Bench Lite and Verified, achieving the highest Best@1 performance among open-source model-based approaches, while maintaining low inference steps and minimal computational overhead.
Notably, SWE-Fixer outperforms several methods based on proprietary models, including those leveraging GPT-4, GPT-4o, and Claude-3-Opus.
By offering a simple yet effective approach to training models for software engineering tasks, SWE-Fixer lowers barriers for the community and fosters further innovation in this domain.

\section*{Limitations}

While our approach achieves high performance on SWE-Bench using open-source LLMs, several limitations must be acknowledged.  
First, the training data scale could be expanded to further improve model performance. Due to computational resource constraints, we are unable to train the model on a significantly larger dataset. Future work can focus on scaling up the training data to enhance performance on the SWE-Bench task.  
Second, a reward model could be introduced for test-time optimization. By constructing negative samples from the training data, a reward model could be trained to evaluate whether a generated code patch successfully resolves an issue. This reward model could then be incorporated into a Best-of-N selection strategy within our pipeline to further refine results.  
Despite these limitations, our method offers a low-cost and effective open-source approach to addressing real-world software engineering problems.

\bibliography{ref}

\begin{thebibliography}{40}
\providecommand{\natexlab}[1]{#1}

\bibitem[{Achiam et~al.(2023)Achiam, Adler, Agarwal, Ahmad, Akkaya, Aleman, Almeida, Altenschmidt, Altman, Anadkat et~al.}]{gpt4}
Josh Achiam, Steven Adler, Sandhini Agarwal, Lama Ahmad, Ilge Akkaya, Florencia~Leoni Aleman, Diogo Almeida, Janko Altenschmidt, Sam Altman, Shyamal Anadkat, et~al. 2023.
\newblock Gpt-4 technical report.
\newblock \emph{arXiv preprint arXiv:2303.08774}.

\bibitem[{{Albert Örwall}(2024)}]{moatless}
{Albert Örwall}. 2024.
\newblock moatless-tools.
\newblock \url{https://github.com/aorwall/moatless-tools}.

\bibitem[{{Anthropic}(2024)}]{claude3_5}
{Anthropic}. 2024.
\newblock Claude 3.5 sonnet.
\newblock \url{https://www.anthropic.com/news/claude-3-5-sonnet}.

\bibitem[{Antoniades et~al.(2024)Antoniades, {\"O}rwall, Zhang, Xie, Goyal, and Wang}]{antoniades2024swe_search}
Antonis Antoniades, Albert {\"O}rwall, Kexun Zhang, Yuxi Xie, Anirudh Goyal, and William Wang. 2024.
\newblock Swe-search: Enhancing software agents with monte carlo tree search and iterative refinement.
\newblock \emph{arXiv preprint arXiv:2410.20285}.

\bibitem[{{AppMap}(2024)}]{appmap2024navie}
{AppMap}. 2024.
\newblock Appmap navie.
\newblock \url{https://appmap.io/blog/2024/06/20/appmap-navie-swe-bench-leader/}.

\bibitem[{Bui et~al.(2021)Bui, Yu, and Jiang}]{bui2021self}
Nghi~DQ Bui, Yijun Yu, and Lingxiao Jiang. 2021.
\newblock Self-supervised contrastive learning for code retrieval and summarization via semantic-preserving transformations.
\newblock In \emph{Proceedings of the 44th International ACM SIGIR Conference on Research and Development in Information Retrieval}, pages 511--521.

\bibitem[{Chen et~al.(2021)Chen, Tworek, Jun, Yuan, de~Oliveira~Pinto, Kaplan, Edwards, Burda, Joseph, Brockman, Ray, Puri, Krueger, Petrov, Khlaaf, Sastry, Mishkin, Chan, Gray, Ryder, Pavlov, Power, Kaiser, Bavarian, Winter, Tillet, Such, Cummings, Plappert, Chantzis, Barnes, Herbert-Voss, Guss, Nichol, Paino, Tezak, Tang, Babuschkin, Balaji, Jain, Saunders, Hesse, Carr, Leike, Achiam, Misra, Morikawa, Radford, Knight, Brundage, Murati, Mayer, Welinder, McGrew, Amodei, McCandlish, Sutskever, and Zaremba}]{chen2021codex_humaneval}
Mark Chen, Jerry Tworek, Heewoo Jun, Qiming Yuan, Henrique~Ponde de~Oliveira~Pinto, Jared Kaplan, Harri Edwards, Yuri Burda, Nicholas Joseph, Greg Brockman, Alex Ray, Raul Puri, Gretchen Krueger, Michael Petrov, Heidy Khlaaf, Girish Sastry, Pamela Mishkin, Brooke Chan, Scott Gray, Nick Ryder, Mikhail Pavlov, Alethea Power, Lukasz Kaiser, Mohammad Bavarian, Clemens Winter, Philippe Tillet, Felipe~Petroski Such, Dave Cummings, Matthias Plappert, Fotios Chantzis, Elizabeth Barnes, Ariel Herbert-Voss, William~Hebgen Guss, Alex Nichol, Alex Paino, Nikolas Tezak, Jie Tang, Igor Babuschkin, Suchir Balaji, Shantanu Jain, William Saunders, Christopher Hesse, Andrew~N. Carr, Jan Leike, Josh Achiam, Vedant Misra, Evan Morikawa, Alec Radford, Matthew Knight, Miles Brundage, Mira Murati, Katie Mayer, Peter Welinder, Bob McGrew, Dario Amodei, Sam McCandlish, Ilya Sutskever, and Wojciech Zaremba. 2021.
\newblock \href {https://arxiv.org/abs/2107.03374} {Evaluating large language models trained on code}.

\bibitem[{Chen et~al.(2024)Chen, Tian, Liu, Ren, and Sun}]{chen2024jumpcoder}
Mouxiang Chen, Hao Tian, Zhongxin Liu, Xiaoxue Ren, and Jianling Sun. 2024.
\newblock Jumpcoder: Go beyond autoregressive coder via online modification.
\newblock \emph{arXiv preprint arXiv:2401.07870}.

\bibitem[{Contributors(2023)}]{2023xtuner}
XTuner Contributors. 2023.
\newblock Xtuner: A toolkit for efficiently fine-tuning llm.
\newblock \url{https://github.com/InternLM/xtuner}.

\bibitem[{Du et~al.(2024)Du, Wen, Zhu, Xie, Ji, Liu, Shi, and Jin}]{du2024generalization}
Xiaohu Du, Ming Wen, Jiahao Zhu, Zifan Xie, Bin Ji, Huijun Liu, Xuanhua Shi, and Hai Jin. 2024.
\newblock Generalization-enhanced code vulnerability detection via multi-task instruction fine-tuning.
\newblock \emph{arXiv preprint arXiv:2406.03718}.

\bibitem[{Gao et~al.(2023)Gao, Zhang, Chen, and Lam}]{gao2023jsontuning}
Chang Gao, Wenxuan Zhang, Guizhen Chen, and Wai Lam. 2023.
\newblock Jsontuning: Towards generalizable, robust, and controllable instruction tuning.
\newblock \emph{arXiv preprint arXiv:2310.02953}.

\bibitem[{Huang et~al.(2023)Huang, Lu, Chen, Wan, and Duan}]{huang2023enhancing}
Baizhou Huang, Shuai Lu, Weizhu Chen, Xiaojun Wan, and Nan Duan. 2023.
\newblock Enhancing large language models in coding through multi-perspective self-consistency.
\newblock \emph{arXiv preprint arXiv:2309.17272}.

\bibitem[{Hui et~al.(2024)Hui, Yang, Cui, Yang, Liu, Zhang, Liu, Zhang, Yu, Dang et~al.}]{hui2024qwen2}
Binyuan Hui, Jian Yang, Zeyu Cui, Jiaxi Yang, Dayiheng Liu, Lei Zhang, Tianyu Liu, Jiajun Zhang, Bowen Yu, Kai Dang, et~al. 2024.
\newblock Qwen2. 5-coder technical report.
\newblock \emph{arXiv preprint arXiv:2409.12186}.

\bibitem[{Jain et~al.(2024)Jain, Han, Gu, Li, Yan, Zhang, Wang, Solar-Lezama, Sen, and Stoica}]{jain2024livecodebench}
Naman Jain, King Han, Alex Gu, Wen-Ding Li, Fanjia Yan, Tianjun Zhang, Sida Wang, Armando Solar-Lezama, Koushik Sen, and Ion Stoica. 2024.
\newblock Livecodebench: Holistic and contamination free evaluation of large language models for code.
\newblock \emph{arXiv preprint arXiv:2403.07974}.

\bibitem[{Jiang et~al.(2024)Jiang, Sablayrolles, Roux, Mensch, Savary, Bamford, Chaplot, Casas, Hanna, Bressand et~al.}]{jiang2024mixtral}
Albert~Q Jiang, Alexandre Sablayrolles, Antoine Roux, Arthur Mensch, Blanche Savary, Chris Bamford, Devendra~Singh Chaplot, Diego de~las Casas, Emma~Bou Hanna, Florian Bressand, et~al. 2024.
\newblock Mixtral of experts.
\newblock \emph{arXiv preprint arXiv:2401.04088}.

\bibitem[{Jiang et~al.(2023)Jiang, Liu, Lutellier, and Tan}]{jiang2023impact}
Nan Jiang, Kevin Liu, Thibaud Lutellier, and Lin Tan. 2023.
\newblock Impact of code language models on automated program repair.
\newblock In \emph{2023 IEEE/ACM 45th International Conference on Software Engineering (ICSE)}, pages 1430--1442. IEEE.

\bibitem[{Jimenez et~al.(2023)Jimenez, Yang, Wettig, Yao, Pei, Press, and Narasimhan}]{jimenez2023swebench}
Carlos~E Jimenez, John Yang, Alexander Wettig, Shunyu Yao, Kexin Pei, Ofir Press, and Karthik Narasimhan. 2023.
\newblock Swe-bench: Can language models resolve real-world github issues?
\newblock \emph{arXiv preprint arXiv:2310.06770}.

\bibitem[{Li et~al.(2024)Li, Zhou, and Shen}]{li2024rewriting}
Haochen Li, Xin Zhou, and Zhiqi Shen. 2024.
\newblock Rewriting the code: A simple method for large language model augmented code search.
\newblock \emph{arXiv preprint arXiv:2401.04514}.

\bibitem[{Li et~al.(2025)Li, Tang, Wang, and Guo}]{li2025patchpilot}
Hongwei Li, Yuheng Tang, Shiqi Wang, and Wenbo Guo. 2025.
\newblock Patchpilot: A stable and cost-efficient agentic patching framework.
\newblock \emph{arXiv preprint arXiv:2502.02747}.

\bibitem[{Lin et~al.(2024)Lin, Wang, Wen, Chen, and Mao}]{lin2024one}
Bo~Lin, Shangwen Wang, Ming Wen, Liqian Chen, and Xiaoguang Mao. 2024.
\newblock One size does not fit all: Multi-granularity patch generation for better automated program repair.
\newblock In \emph{Proceedings of the 33rd ACM SIGSOFT International Symposium on Software Testing and Analysis}, pages 1554--1566.

\bibitem[{Liu et~al.(2024{\natexlab{a}})Liu, Xia, Wang, and Zhang}]{liu2024your}
Jiawei Liu, Chunqiu~Steven Xia, Yuyao Wang, and Lingming Zhang. 2024{\natexlab{a}}.
\newblock Is your code generated by chatgpt really correct? rigorous evaluation of large language models for code generation.
\newblock \emph{Advances in Neural Information Processing Systems}, 36.

\bibitem[{Liu et~al.(2024{\natexlab{b}})Liu, Lan, Hu, Liu, Zhang, Wang, Shieh, and Zhou}]{liu2024codexgraph}
Xiangyan Liu, Bo~Lan, Zhiyuan Hu, Yang Liu, Zhicheng Zhang, Fei Wang, Michael Shieh, and Wenmeng Zhou. 2024{\natexlab{b}}.
\newblock Codexgraph: Bridging large language models and code repositories via code graph databases.
\newblock \emph{arXiv preprint arXiv:2408.03910}.

\bibitem[{Ma et~al.(2024)Ma, Cao, Cao, Zhang, Chen, Liu, Liu, Li, Huang, and Li}]{ma2024lingma}
Yingwei Ma, Rongyu Cao, Yongchang Cao, Yue Zhang, Jue Chen, Yibo Liu, Yuchen Liu, Binhua Li, Fei Huang, and Yongbin Li. 2024.
\newblock Lingma swe-gpt: An open development-process-centric language model for automated software improvement.
\newblock \emph{arXiv preprint arXiv:2411.00622}.

\bibitem[{Martinez-Gil(2024)}]{martinez2024improving}
Jorge Martinez-Gil. 2024.
\newblock Improving source code similarity detection through graphcodebert and integration of additional features.
\newblock \emph{arXiv preprint arXiv:2408.08903}.

\bibitem[{Ouyang et~al.(2024)Ouyang, Yu, Ma, Xiao, Zhang, Jia, Han, Zhang, and Yu}]{ouyang2024repograph}
Siru Ouyang, Wenhao Yu, Kaixin Ma, Zilin Xiao, Zhihan Zhang, Mengzhao Jia, Jiawei Han, Hongming Zhang, and Dong Yu. 2024.
\newblock Repograph: Enhancing ai software engineering with repository-level code graph.
\newblock \emph{arXiv preprint arXiv:2410.14684}.

\bibitem[{Pan et~al.(2024)Pan, Wang, Neubig, Jaitly, Ji, Suhr, and Zhang}]{pan2024swegym}
Jiayi Pan, Xingyao Wang, Graham Neubig, Navdeep Jaitly, Heng Ji, Alane Suhr, and Yizhe Zhang. 2024.
\newblock Training software engineering agents and verifiers with {SWE-Gym}.

\bibitem[{Qin et~al.(2024)Qin, Wang, Lou, Dong, Wang, Li, and Mao}]{qin2024agentfl}
Yihao Qin, Shangwen Wang, Yiling Lou, Jinhao Dong, Kaixin Wang, Xiaoling Li, and Xiaoguang Mao. 2024.
\newblock Agentfl: Scaling llm-based fault localization to project-level context.
\newblock \emph{arXiv preprint arXiv:2403.16362}.

\bibitem[{Qwen et~al.(2024)Qwen, :, Yang, Yang, Zhang, Hui, Zheng, Yu, Li, Liu, Huang, Wei, Lin, Yang, Tu, Zhang, Yang, Yang, Zhou, Lin, Dang, Lu, Bao, Yang, Yu, Li, Xue, Zhang, Zhu, Men, Lin, Li, Xia, Ren, Ren, Fan, Su, Zhang, Wan, Liu, Cui, Zhang, and Qiu}]{qwen2024qwen25technicalreport}
Qwen, :, An~Yang, Baosong Yang, Beichen Zhang, Binyuan Hui, Bo~Zheng, Bowen Yu, Chengyuan Li, Dayiheng Liu, Fei Huang, Haoran Wei, Huan Lin, Jian Yang, Jianhong Tu, Jianwei Zhang, Jianxin Yang, Jiaxi Yang, Jingren Zhou, Junyang Lin, Kai Dang, Keming Lu, Keqin Bao, Kexin Yang, Le~Yu, Mei Li, Mingfeng Xue, Pei Zhang, Qin Zhu, Rui Men, Runji Lin, Tianhao Li, Tingyu Xia, Xingzhang Ren, Xuancheng Ren, Yang Fan, Yang Su, Yichang Zhang, Yu~Wan, Yuqiong Liu, Zeyu Cui, Zhenru Zhang, and Zihan Qiu. 2024.
\newblock \href {https://arxiv.org/abs/2412.15115} {Qwen2.5 technical report}.
\newblock \emph{Preprint}, arXiv:2412.15115.

\bibitem[{Robertson et~al.(2009)Robertson, Zaragoza et~al.}]{robertson2009probabilistic}
Stephen Robertson, Hugo Zaragoza, et~al. 2009.
\newblock The probabilistic relevance framework: Bm25 and beyond.
\newblock \emph{Foundations and Trends{\textregistered} in Information Retrieval}, 3(4):333--389.

\bibitem[{Saieva et~al.(2023)Saieva, Chakraborty, and Kaiser}]{saieva2023reinforest}
Anthony Saieva, Saikat Chakraborty, and Gail Kaiser. 2023.
\newblock Reinforest: Reinforcing semantic code similarity for cross-lingual code search models.
\newblock \emph{arXiv preprint arXiv:2305.03843}.

\bibitem[{Team et~al.(2023)Team, Anil, Borgeaud, Wu, Alayrac, Yu, Soricut, Schalkwyk, Dai, Hauth et~al.}]{team2023gemini}
Gemini Team, Rohan Anil, Sebastian Borgeaud, Yonghui Wu, Jean-Baptiste Alayrac, Jiahui Yu, Radu Soricut, Johan Schalkwyk, Andrew~M Dai, Anja Hauth, et~al. 2023.
\newblock Gemini: a family of highly capable multimodal models.
\newblock \emph{arXiv preprint arXiv:2312.11805}.

\bibitem[{Wang et~al.(2024{\natexlab{a}})Wang, Cassano, Wu, Bai, Song, Nath, Han, Hendryx, Yue, and Zhang}]{wang2024planning}
Evan Wang, Federico Cassano, Catherine Wu, Yunfeng Bai, Will Song, Vaskar Nath, Ziwen Han, Sean Hendryx, Summer Yue, and Hugh Zhang. 2024{\natexlab{a}}.
\newblock Planning in natural language improves llm search for code generation.
\newblock \emph{arXiv preprint arXiv:2409.03733}.

\bibitem[{Wang et~al.(2024{\natexlab{b}})Wang, Li, Song, Xu, Tang, Zhuge, Pan, Song, Li, Singh et~al.}]{wang2024opendevin}
Xingyao Wang, Boxuan Li, Yufan Song, Frank~F Xu, Xiangru Tang, Mingchen Zhuge, Jiayi Pan, Yueqi Song, Bowen Li, Jaskirat Singh, et~al. 2024{\natexlab{b}}.
\newblock Opendevin: An open platform for ai software developers as generalist agents.
\newblock \emph{arXiv preprint arXiv:2407.16741}.

\bibitem[{Xia et~al.(2024)Xia, Deng, Dunn, and Zhang}]{xia2024agentless}
Chunqiu~Steven Xia, Yinlin Deng, Soren Dunn, and Lingming Zhang. 2024.
\newblock Agentless: Demystifying llm-based software engineering agents.
\newblock \emph{arXiv preprint arXiv:2407.01489}.

\bibitem[{Xia et~al.(2022)Xia, Wei, and Zhang}]{xia2022practical}
Chunqiu~Steven Xia, Yuxiang Wei, and Lingming Zhang. 2022.
\newblock Practical program repair in the era of large pre-trained language models.
\newblock \emph{arXiv preprint arXiv:2210.14179}.

\bibitem[{Yang et~al.(2024)Yang, Jimenez, Wettig, Lieret, Yao, Narasimhan, and Press}]{yang2024sweagent}
John Yang, Carlos~E Jimenez, Alexander Wettig, Kilian Lieret, Shunyu Yao, Karthik Narasimhan, and Ofir Press. 2024.
\newblock Swe-agent: Agent-computer interfaces enable automated software engineering.
\newblock \emph{arXiv preprint arXiv:2405.15793}.

\bibitem[{Zelikman et~al.(2022)Zelikman, Wu, Mu, and Goodman}]{zelikman2022star}
Eric Zelikman, Yuhuai Wu, Jesse Mu, and Noah Goodman. 2022.
\newblock Star: Bootstrapping reasoning with reasoning.
\newblock \emph{Advances in Neural Information Processing Systems}, 35:15476--15488.

\bibitem[{Zhang et~al.(2024)Zhang, Ruan, Fan, and Roychoudhury}]{zhang2024autocoderover}
Yuntong Zhang, Haifeng Ruan, Zhiyu Fan, and Abhik Roychoudhury. 2024.
\newblock Autocoderover: Autonomous program improvement.
\newblock In \emph{Proceedings of the 33rd ACM SIGSOFT International Symposium on Software Testing and Analysis}, pages 1592--1604.

\bibitem[{Zheng et~al.(2023)Zheng, Yuan, Zhang, Yang, and Kong}]{zheng2023self}
Lin Zheng, Jianbo Yuan, Zhi Zhang, Hongxia Yang, and Lingpeng Kong. 2023.
\newblock Self-infilling code generation.
\newblock In \emph{Forty-first International Conference on Machine Learning}.

\bibitem[{Zhu et~al.(2024)Zhu, Guo, Shao, Yang, Wang, Xu, Wu, Li, Gao, Ma et~al.}]{zhu2024deepseekcoderv2}
Qihao Zhu, Daya Guo, Zhihong Shao, Dejian Yang, Peiyi Wang, Runxin Xu, Y~Wu, Yukun Li, Huazuo Gao, Shirong Ma, et~al. 2024.
\newblock Deepseek-coder-v2: Breaking the barrier of closed-source models in code intelligence.
\newblock \emph{arXiv preprint arXiv:2406.11931}.

\end{thebibliography}

\appendix

\section{Training Data Collection}
\label{sec:trainging_data_collection}
In line with the data collection methodology outlined in SWE-Bench \citep{jimenez2023swebench}, we gather high-quality issues, pull requests, and codebases from GitHub repositories for training purposes. Subsequently, we apply a filtering process to enhance data quality and eliminate overly complex training examples.

\paragraph{Repository Collection}
We use Github REST API to crawl a list of initial repositories. 
We select Python repositories that have more than 100 pull requests (PRs).
In the official SWE-Bench code base, an instance extraction script is used to crawl SWE-Bench-style instances, each corresponding to an issue with a PR and a set of potential unit tests.
We find that this script relies on string match \footnote{{\scriptsize \url{https://docs.github.com/en/issues/tracking-your-work-with-issues/using-issues/linking-a-pull-request-to-an-issue\#linking-a-pull-request-to-an-issue-using-a-keyword}}} to identify valid issue PR pairs. Therefore, it has a large chance to miss out many instances.
We instead rely on GitHub events\footnote{\url{https://docs.github.com/en/rest/using-the-rest-api/issue-event-types?apiVersion=2022-11-28}} to conduct the crawling job. 
Eventually, for the raw data, we collect 2.3K repositories and 331K instances, where repositories within SWE-Bench have been excluded.

\paragraph{Data Statistics}

\begin{figure}[h!]
    \centering
    \begin{subfigure}{0.35\textwidth}
        \setlength{\abovecaptionskip}{10pt}
        \centering
        
        \includegraphics[width=\textwidth]{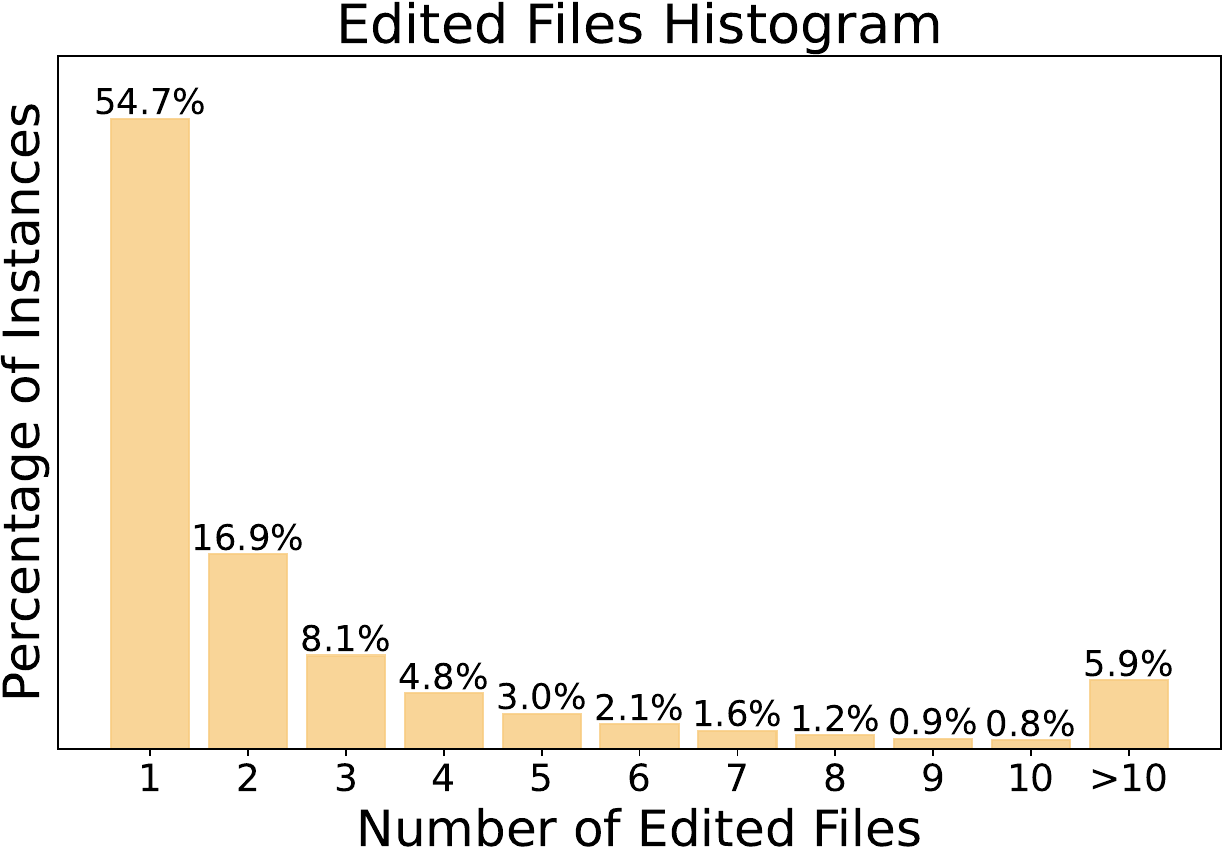} %
        \caption{}
        \label{fig:14w_edited_files_statistics}
    \end{subfigure}   

    \begin{subfigure}{0.35\textwidth}
        \adjustbox{trim=0mm 0mm 0mm 0mm}{\includegraphics[width=\textwidth]{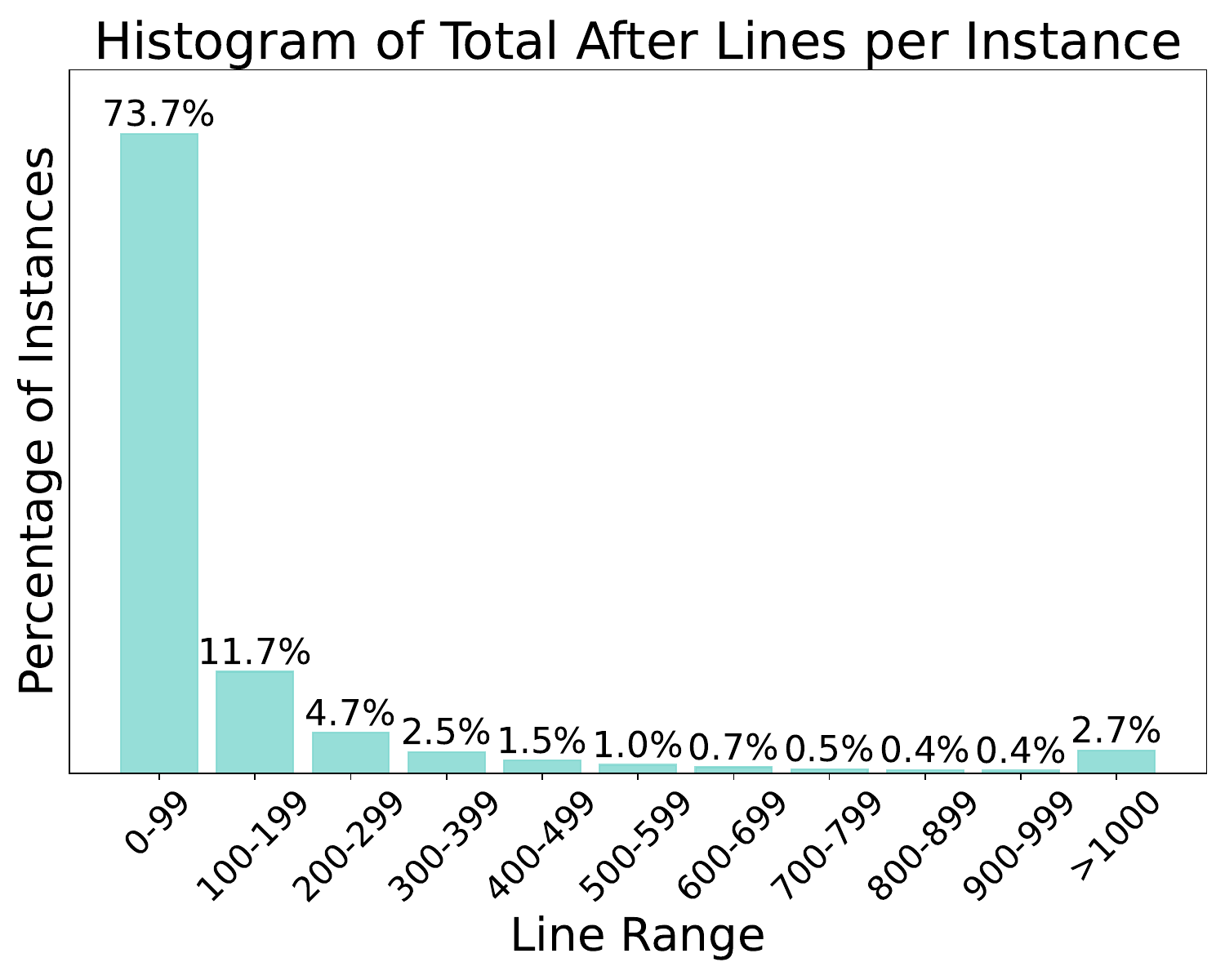}}

        \caption{}
        \label{fig:14w_modified_lines_distribution}
    \end{subfigure}
    \begin{subfigure}{0.35\textwidth}
        \setlength{\abovecaptionskip}{10pt} %
        \centering
        \includegraphics[width=\textwidth]{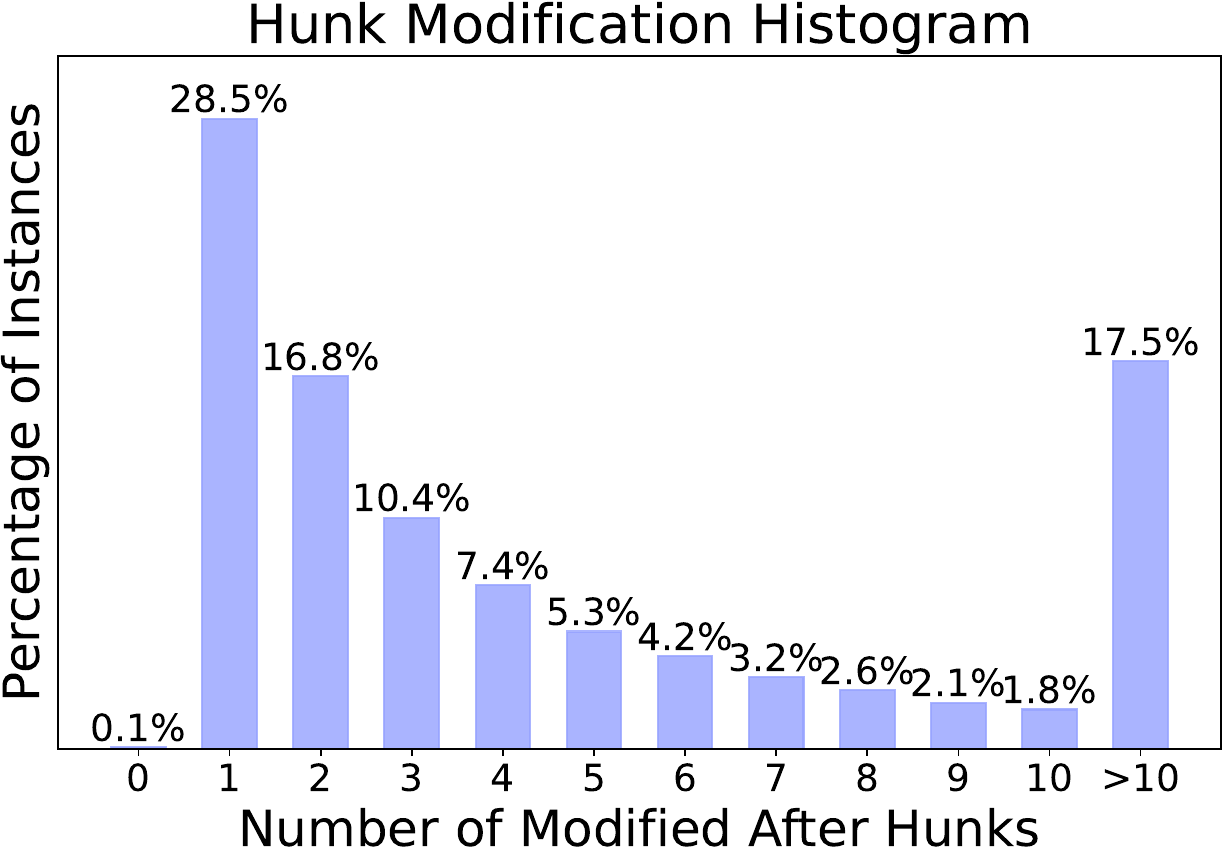} %
        
        \caption{}
        \label{fig:14w_modified_hunks_statistics}
    \end{subfigure}
    \caption{We sample a subset of the crawled raw data to analyze the statistics on real-world data. (a) The histogram of the number of edited files. (b) The distribution of the number of edited code lines. (c) The histogram shows the number of edited code hunks. 
    }
    \label{fig:14w_data_statistic}
\end{figure}

As shown in Figure \ref{fig:14w_edited_files_statistics}, most real-world instances involve a small number of edited files. Specifically, 54.7\% of instances involve modifications to only one file, and nearly 80\% involve no more than three files. 
Figure \ref{fig:14w_modified_lines_distribution} shows the distribution of modified lines, with 73.7\% of instances involving modifications to 0–99 lines and over 85\% involving up to 200 lines. 
Similarly, Figure \ref{fig:14w_modified_hunks_statistics} reveals that instances involving more modification regions (hunks) occur less frequently.

\paragraph{Data Filtering}
To address the challenges posed by the messy nature of real-world data, we apply instance-level data filtering to ensure higher data quality. 
For the sake of efficiency, we sample a subset of 140K instances, parse the oracle code patches and discard instances whose patches cannot be parsed. 
Then we filter out instances where the number of edited files (excluding test files) is more than three. 
This decision is based on our observation in the data statistics and editing more than three files introduces excessive complexity, hindering effective model training.
This process results in a train set of 110K instances, \trainset.

\begin{table}[t]
\centering
\caption{Impact of retrieval training methods of the 7B model on the overall pipeline performance of SWE-Bench Lite, using the 72B editor model. 
Both the retrieval and editor models are trained on \trainsmall. 
$^\dag$: In this setup, the retriever is trained to additionally retrieve class and function names, where the 72B editor model is also specifically trained to incorporate the input change.
}
\small
\setlength{\tabcolsep}{4pt}
\scalebox{0.95}{
\begin{tabular*}{\linewidth}{@{\extracolsep{\fill}}lcc}
\toprule
\textbf{Method} & \textbf{Pre/Recall(\%)} & \textbf{Resolve(\%)} \\ 
\midrule
Base setting  & \textbf{65.4}/67.3 & \textbf{16.3}\\
~~{ - Also retrieve Cls\&Func}$^\dag$ & 54.0/56.0 & 15.7~\textcolor{red}{($\downarrow$~0.6)} \\ 
\bottomrule
\end{tabular*} 
}
\label{tab:no_class_exp}
\end{table}

\section{Post-processing}
\label{sec:post_pocessing}
To ensure that the model-generated outputs meet the task requirements, we implement a post-processing procedure with a unified resampling strategy for both the retrieval and code editing tasks to ensure output validity. The sampling temperature is initially set to 0 for deterministic outputs. 
If the output fails or is invalid, the temperature is adjusted to 0.7 for further attempts. To prevent endless retries, the maximum number of attempts is set to 5.
For the retrieval task, outputs are checked for JSON syntax validation, which means, triggering resampling if invalid. For the code editing task, we similarly require the model-generated results to initially pass JSON validation. Once validated, the model-generated modifications are applied to the original file. If the original code snippet cannot be located, the result is deemed invalid. Additionally, the modification is considered invalid if the modified code fails to pass syntax checks or if it does not pass existing test files in the repository.
This post-processing procedure effectively improves the correctness of the model-generated outputs in terms of format and content, ensuring that they better meet the requirements of real-world applications.

\section{P2P Filtering}
\label{sec:p2p}
Pass-to-Pass (P2P) filtering refers to using regression tests from each code repository to validate the correctness of generated patches. Specifically, a P2P test is a test case that passes both before and after applying the gold patch, and can be considered a check for the correctness of unrelated functionality in the repository. SWE-Bench provides a list of such tests for each issue. Following the setup of earlier works such as Agentless v1~\cite{xia2024agentless} and Aider, in some experiments, we apply P2P filtering in our inference stage.
Concretely, we apply the generated patch to the repository and verify whether it breaks any P2P tests. If the patch passes all P2P tests, it is retained; otherwise, we resample a new patch. 
We first use greedy sampling to generate the patch. If the patch fails, a temperature of 0.7 is applied to generate more creative outputs.
On instances where the patch ultimately passes, the average number of generation attempts is 1.15 on SWE-Bench Lite and 6.73 on SWE-Bench Verified.
Given the ongoing community discussion on the SWE-Bench GitHub repository regarding whether P2P filtering should be used during inference, we report both P2P-filtered and non-filtered results in our paper for transparency and comparison.

\section{Model Calls Calculation for SWE-Seach}
\label{sec:swe_search_step_estimation}
Since SWE-Search has not released its execution trajectory, we can estimate the required inference steps based on the methodology section. The original paper states: "In SWE-Search, we limit each node to a maximum of three expansions and cap the total search iterations at 100." This implies that generating an answer requires at least 100 steps. Additionally, since each node needs to be evaluated by the value model, there are also at least 100 evaluations, leading to a minimum of 200 inference steps. However, the actual number of inferences is likely much higher because a new iteration involves generating more than just one new node.

\section{Ablation on More Fine-Grained Retrieval Strategies}
\label{sec:more_finegained_exp}
We also experiment with more fine-grained retrieval strategies during the retrieval stage to determine if additional details could enhance overall performance in our framework. However, this extra information increases the complexity of the retrieval task and ultimately reduces the overall pipeline performance when class and function names are retrieved (see Table \ref{tab:no_class_exp}).

\clearpage
\newpage
\section{CoT Generation Prompt}
\label{sec:cot_generation_prompt}

\begin{center}
    \begin{minipage}{0.9\textwidth}  %
        \begin{customprompt}[]{Cot Generation System Prompt}
        You are an expert software engineer and seasoned code reviewer, specializing in bug localization and code optimization within real-world code repositories. Your role is to meticulously analyze code and provide clear, logical reasoning to guide the resolution of issues within the codebase.
        
        In this role, you focus on the precision and effectiveness of the problem-solving process. Your expertise includes understanding complex code structures and accurately mapping issues to the specific parts of the code requiring modification. You excel at breaking down the reasoning process into coherent, easy-to-follow steps that lead to efficient and accurate code fixes.
        
        In this task, we are training a model to generate code modifications for resolving issues within real-world codebases. For this, we have the issue description, the codebase, and the corresponding oracle code modifications. Your task is to generate detailed reasoning to aid in collecting high-quality training data. Your reasoning process must be thorough, evidence-based, and strictly adhere to the provided issue.
        
        Although oracle code modifications are available, **you should simulate reasoning independently—as if you are identifying the necessary files and changes without prior knowledge of those modifications**. Avoid statements like "The edited code makes sense because…" that imply direct knowledge of the oracle modifications.
        \end{customprompt}
    \end{minipage}
\end{center}
\clearpage
\newpage

\begin{figure*}[t]
    \centering
    \begin{minipage}{0.9\textwidth}  %
\begin{customprompt}{Cot Generation User Prompt}
\# Issue Statement:
\{problem\_statement\}

\# File Content to be Modified:
You are provided with the files that require modification to resolve the issue. This includes the full file content. You should identify the code snippets to be modified based on the issue and the file content.
\{content\}

\# Oracle Code Modifications:
\{target\}

\# Task Objective:
Your objective is to develop a clear and logical reasoning process that guides the modification of the code snippets based on the issue. The reasoning should explain the relationship between the issue and each code snippet, and why the modifications are necessary.

\# Reasoning Process Guidelines:
The reasoning process should generally include the following steps. You may adjust these steps as needed for clarity and accuracy:

1. **Issue Analysis**:
   - Begin by **clearly articulating the issue**. Provide a comprehensive explanation of why this issue is significant, highlighting the specific challenges or obstacles that must be addressed. Identify the key requirements or objectives necessary for resolving the issue, ensuring that all aspects of the issue are thoroughly examined and understood.

2. **Task Decomposition*:
   - Break down the overall issue into **smaller, manageable sub-tasks**. Explain the purpose of each sub-task and its significance in solving the issue. Ensure that sub-tasks are logically ordered and clearly connected.

3. **Code Localization and Editing**:
   - First, for each sub-task, identify the relevant **code snippet** by providing the file path and referring to the specific part of the code related to that sub-task. Next, give a detailed explanation of how this code snippet is connected to the sub-task, explain how the code should be edited to resolve the issue and justify why these changes are necessary. Finally, provide the edited code based on the explanation.
   - Ensure that the final output for this part MATCHES the provided oracle modifications EXACTLY.

\# General Requirements:

1. **Clear and Evidence-Based Reasoning**: Provide clear and precise reasoning for each step, strictly based on the provided issue and code without inferring information not explicitly stated.
2. **Comprehensive and Concise**: Address all relevant aspects of the issue comprehensively while being concise. Justify the exclusion of any sections that are not relevant.
3. **Detailed Guidance**: Ensure the reasoning steps are detailed enough to allow someone unfamiliar with the solution to infer and implement the necessary code modifications.
4. **Faithfulness**: Ensure that your final output for the code modifications MATCHES the provided oracle modifications EXACTLY.
5. **Neutral Perspective**: Approach the issue as if you do not know the correct answer in advance. Avoid language that implies prior knowledge of the correct modifications.

\# Format Requirements:

1. **File path**: Always mention the file path when referring to a code snippet (including class or function names).
2. **Reasoning Process Format**: Use markdown to present your reasoning process. Clearly define each step and ensure logical connections between them.
3. **Code Snippet**: You must include **line numbers** when referring to the original code for context and outputing `code\_snippet\_to\_be\_modified`. However, do **not include line numbers** in your editing suggestions.

Please ensure your response is clearly formatted and provides enough detail to justify why each code section was selected for modification and how it should be edited.

\end{customprompt}

    \end{minipage}
\end{figure*}

\clearpage
\newpage

\section{File Documentation}
\label{sec:file_skeleton}
\begin{center}
    \begin{figure}[h]
    \includegraphics[width=\linewidth]{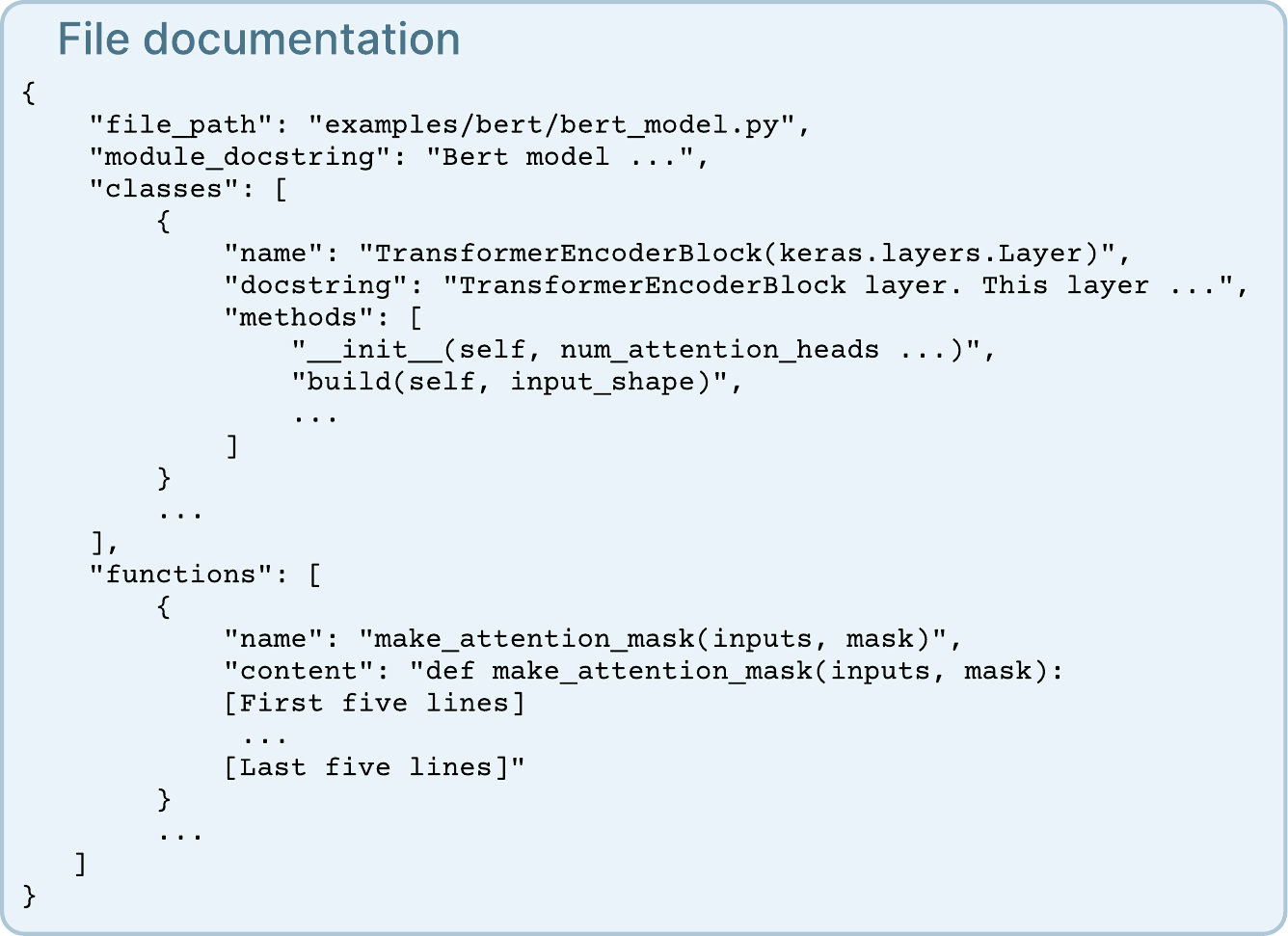}
    \caption{Example of a file documentation. The documentation includes the relative file path and module docstring (if available). It also contains class names, their associated docstrings, and all method names. For functions, only the name and the first/last five lines of code are included. 
    }
    \label{fig:file_skeleton}
    \end{figure}
    
\end{center}
\clearpage
\newpage

\section{Edit Output Patch Example}
\label{sec:our_edit_output_example}
\begin{center}
    \begin{minipage}{\linewidth}
    \includegraphics{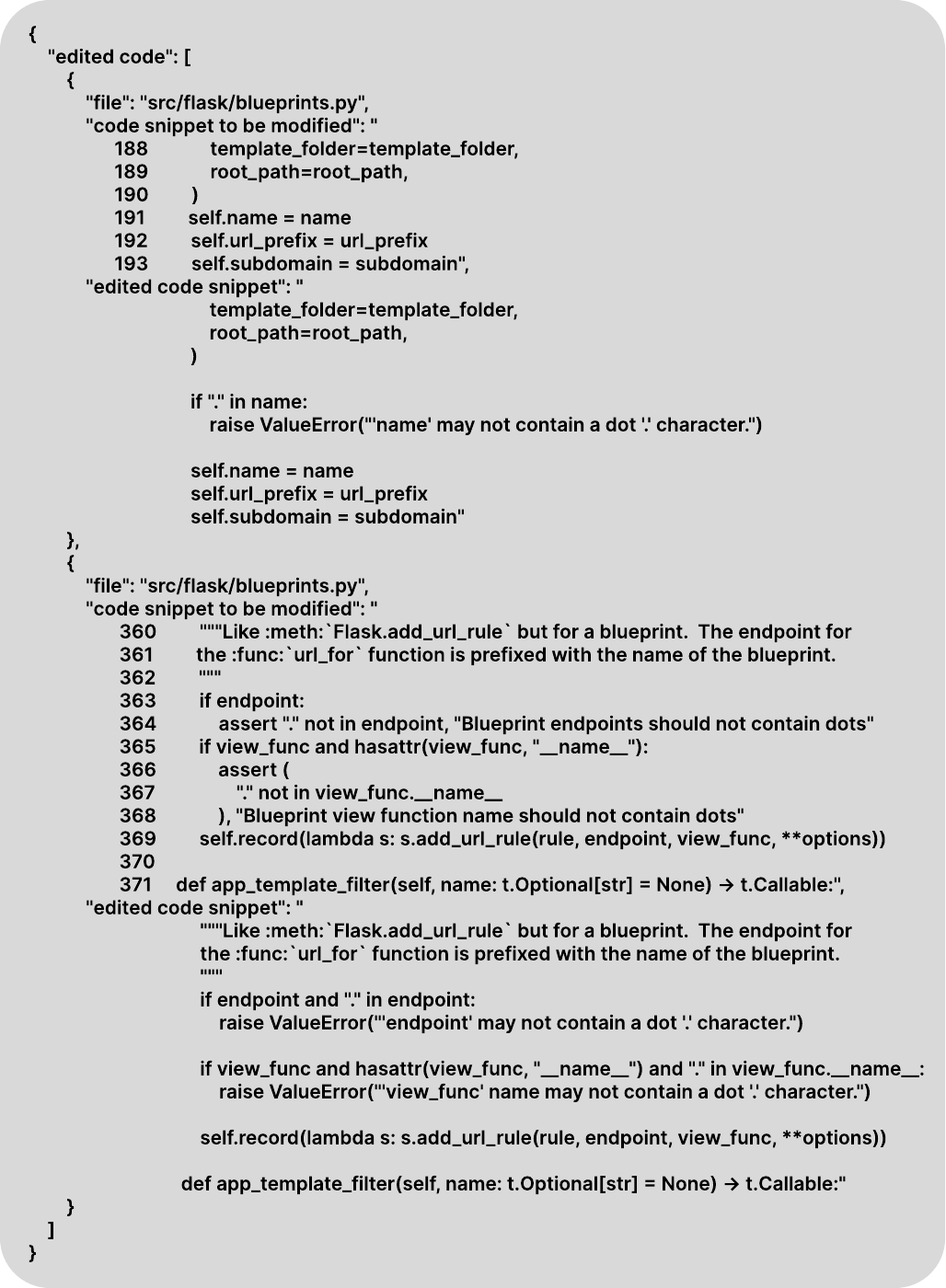}
    \label{fig:our_edit_output_example}
    \end{minipage}
\end{center}

\end{document}